\documentclass[10pt,twocolumn,letterpaper]{article}

\usepackage[pagenumbers]{cvpr} 
\usepackage{amsmath}

\usepackage[dvipsnames]{xcolor}

\definecolor{cvprblue}{rgb}{0.21,0.49,0.74}
\usepackage[pagebackref,breaklinks,colorlinks,citecolor=cvprblue]{hyperref}
\usepackage{pifont}
\usepackage{multirow}
\usepackage{arydshln} 
\usepackage{makecell}
\usepackage{wrapfig}

\def\paperID{14351	
} 
\def\confName{CVPR}
\def\confYear{2024}

\title{Doubly Abductive Counterfactual Inference for Text-based Image Editing}

\author{
Xue Song\textsuperscript{1}, Jiequan Cui\textsuperscript{3}, Hanwang Zhang\textsuperscript{2,3}, Jingjing Chen\textsuperscript{1}\thanks{Corresponding author},
Richang Hong\textsuperscript{4}, Yu-Gang Jiang\textsuperscript{1}\\
{\small \textsuperscript{1}Shanghai Key Lab of Intell. Info. Processing, School of CS, Fudan University} \\
{\small \textsuperscript{2}Skywork AI \qquad \textsuperscript{3}Nanyang Technological University \qquad \textsuperscript{4}Hefei University of Technology} \\
{\tt\footnotesize \{xsong18, chenjingjing, ygj\}@fudan.edu.cn, hanwangzhang@ntu.edu.sg, \{jiequancui, hongrc.hfut\}@gmail.com}}

\begin{document}
\maketitle
\begin{abstract}
We study text-based image editing (TBIE) of a single image by counterfactual inference because it is an elegant formulation to precisely address the requirement: the edited image should retain the fidelity of the original one. Through the lens of the formulation, we find that the crux of TBIE is that existing techniques hardly achieve a good trade-off between editability and fidelity, mainly due to the overfitting of the single-image fine-tuning.  To this end, we propose a Doubly Abductive Counterfactual inference framework (DAC). We first parameterize an exogenous variable as a UNet LoRA, whose abduction can encode all the image details. Second, we abduct another exogenous variable parameterized by a text encoder LoRA, which recovers the lost editability caused by the overfitted first abduction. Thanks to the second abduction, which exclusively encodes the visual transition from post-edit to pre-edit, its inversion---subtracting the LoRA---effectively reverts pre-edit back to post-edit, thereby accomplishing the edit. Through extensive experiments, our DAC achieves a good trade-off between editability and fidelity. Thus, we can support a wide spectrum of user editing intents, including addition, removal, manipulation, replacement, style transfer, and facial change, which are extensively validated in both qualitative and quantitative evaluations. Codes are in \url{https://github.com/xuesong39/DAC}.
\end{abstract}
    
\section{Introduction}
\label{sec:introduction}

\begin{figure}[t]
    \centering
    \subfloat[Source image]{ \includegraphics[width=0.15\textwidth]{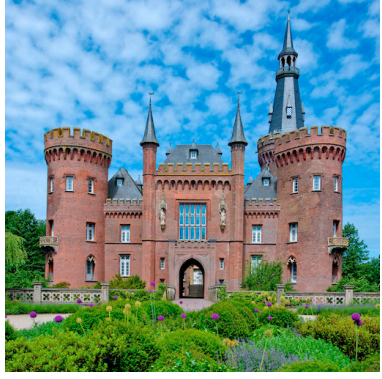}}     
    \subfloat[Edited image]{ \includegraphics[width=0.15\textwidth]{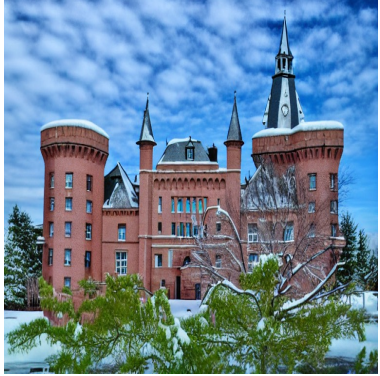}}  
    \subfloat[Edited image]{ \includegraphics[width=0.15\textwidth]{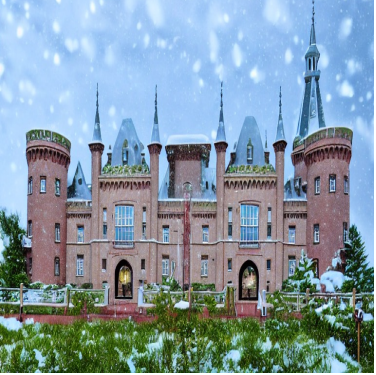}} 
    \vspace{-0.1in}
    \caption{
    Illustration of the TBIE task. (a): source image $I$. (b) and (c): edited images according to the target prompt ``a castle covered by snow''. TBIE considers (b) to be better than (c).
    }
\label{fig:illustration_image_editing}
\vspace{-3mm}
\end{figure}

Text-based image editing (TBIE) modifies a user-uploaded real image to match a textual prompt while keeping minimal visual changes---the fidelity of the original image.  
As shown in Figure~\ref{fig:illustration_image_editing}, the source image $I$ in (a) is edited with the prompt ``I want the castle covered by snow''. We consider the edited image $I'$ in (b) to be better than that in (c) because the former keeps a better structure of the castle, leading to minimal changes to the source image. Without loss of generality\footnote[1]{Any LLM with proper instruction tuning or in-context learning can interpret the user intent into $P$ and $P'$. We have deliberately excluded this module from our formulation.}, we denote the prompt into two sub-prompts $P$ and $P'$, where $P$ describes the image content of user's editing intent and $P'$ describes it after editing. For example, $P$ is ``a castle'' and $P'$ is ``a castle covered by snow''. 

\begin{figure}[t]
    \centering
    \subfloat[Abduction]{ \includegraphics[width=0.18\textwidth]{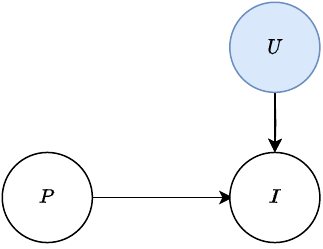}}    
    \hspace{+0.02in}
    \subfloat[Action \& Prediction]{ \includegraphics[width=0.18\textwidth]{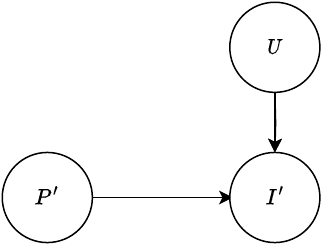}}  
    \vspace{-0.10in}
    \caption{
        Counterfactual inference framework for TBIE.
    }
\label{fig:counterfactual_inference}
\vspace{-6mm}
\end{figure}

TBIE is a challenging task as it is inherently zero-shot: a source image $I$ and a prompt ($P$, $P'$) are the only input and there is no ground-truth image for the target image $I'$.
Fortunately, thanks to the large-scale text-to-image generative models, \eg, DALL-E~\cite{ramesh2022hierarchical}, Imagen~\cite{saharia2022photorealistic}, and Stable Diffusion~\cite{rombach2022high}, language embeddings and visual features are well-aligned. So, they provide a channel to modify images via natural language. However, the editing efficacy of existing methods is still far from satisfactory, for example, they can only support limited edits like style transfer~\cite{kim2022diffusionclip}, add/remove objects~\cite{avrahami2022blended}; do not support user-uploaded images~\cite{hertz2022prompt}, or require extra supervision~\cite{ruiz2023dreambooth} and spatial masks to localize where to edit~\cite{avrahami2022blended}.

Yet, there is no theory that explains why TBIE is challenging, or why existing methods sometimes succeed or fail. Such an absence will undoubtedly hinder progress in this field.
To this end, as illustrated in Figure~\ref{fig:counterfactual_inference}, we formulate TBIE as a counterfactual inference problem~\cite{pearl2016causal} based on text-conditional diffusion models, \eg, we use Stable Diffusion~\cite{rombach2022high} in this paper.

\begin{figure}[t]
    \centering
    \includegraphics[width=0.5\textwidth]{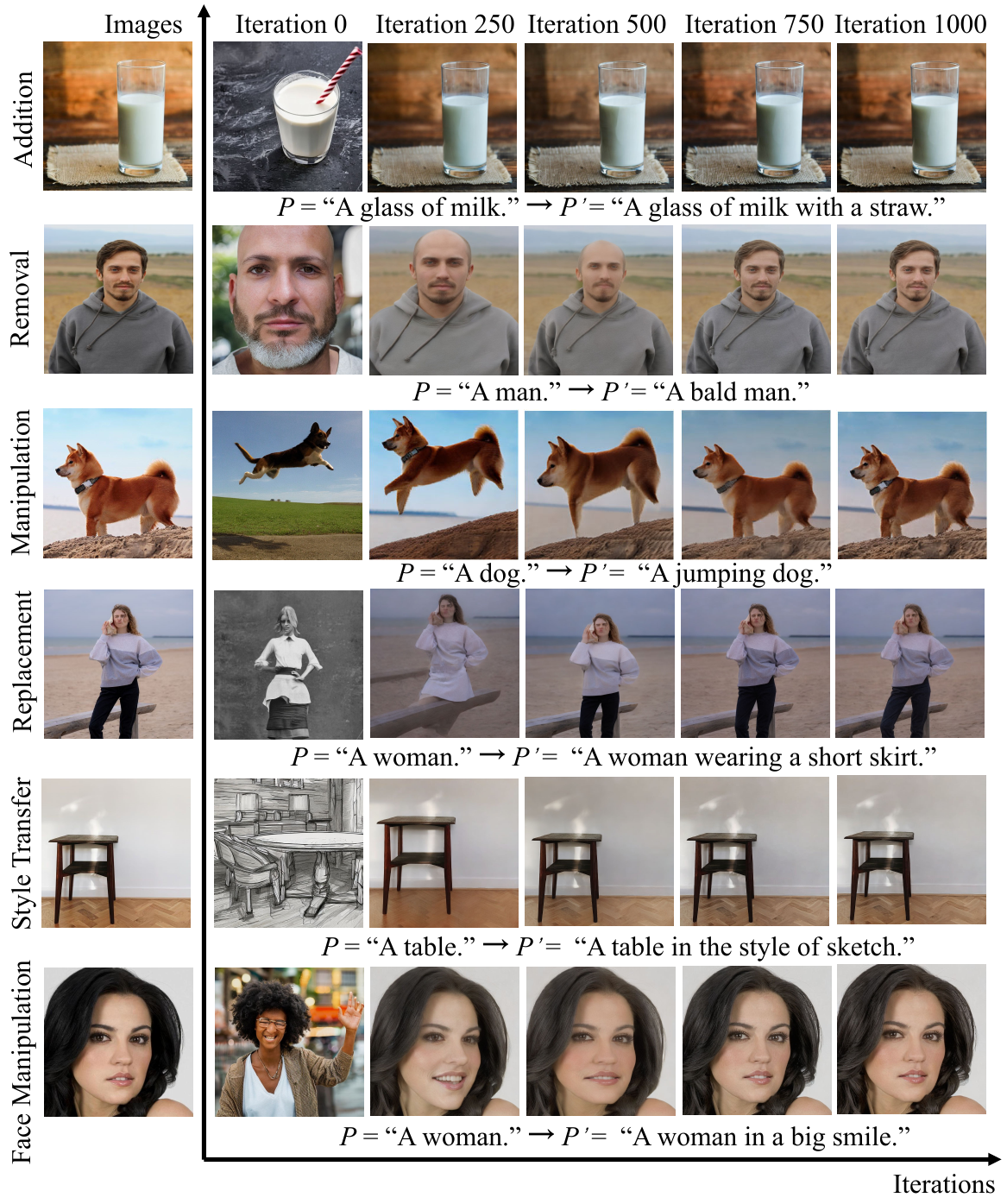} 
    \vspace{-6.5mm}
    \caption{
        The editability of counterfactual $I' = G(P', U)$ decreases when the abductive iteration of $\arg\min_{U} \|G(P, U)- I\|$ increases.
    }
 \vspace{-6.5mm}
\label{fig:editability_interations}
\end{figure}

\noindent\textbf{Why Counterfactual?}
Counterfactual inference can define the ``minimal visual change'' requirement formally. As prompt $P$ describes the existing contents in source image $I$, the generative model $G$ should be able to generate $I$ based on $P$. However, $G$ is usually probabilistic, \ie, only $P$ is not enough to control $G$ to generate an image exactly the same as $I$, thus we need an unknown exogenous variable $U$ to remove the uncertainty:
\begin{equation}
\textrm{Fact}: I = G(P, U). \label{eq:1}
\end{equation}
Therefore, the ``minimal visual change'' in TBIE can be formulated as the following counterfactual:
\begin{equation}
\textrm{Counterfactual}: I' = G(P', U),
\end{equation}
where $U$ is abducted from Eq.~\eqref{eq:1} by $\arg\min_{U} \|G(P, U)- I\|$ to ensure that the edited image $I'$ preserves most of the visual content of $I$ while incorporating the influence of $P'$.

\noindent\textbf{Why Challenging?}
The abduction of $U$ is inevitably ill-posed, \ie, $U$ overfits to the particular $P$ and $I$. As a result, $G(\cdot, U)$ may ruin the pre-trained prior distribution and fail to comprehend $P'$. As shown in Figure~\ref{fig:editability_interations},  as the number of iterations of $\arg\min_{U} \|G(P, U)- I\|$ increases, $G(P', U)$ generates $I'$ more similar to $I$, but at the same time, the editability of $G(P', U)$ is decreasing. However, it is elusive to find a good $U$ that balances the trade-off between editability and fidelity.
Thanks to the counterfactual framework, we conjecture that the success or failure of existing TBIE methods is primarily attributed to the trade-off (Section~\ref{sec:relatedwork}).

\begin{figure}[t]
    \subfloat[Abduction 1]{ \includegraphics[width=0.15\textwidth]{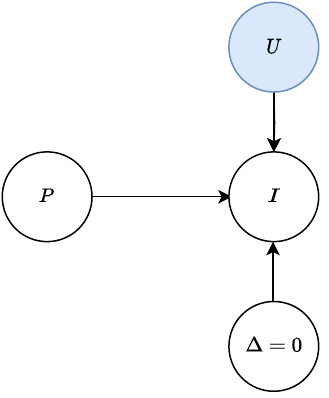}} 
    \hspace{+0.01in}
    \subfloat[Abduction 2]{ \includegraphics[width=0.15\textwidth]{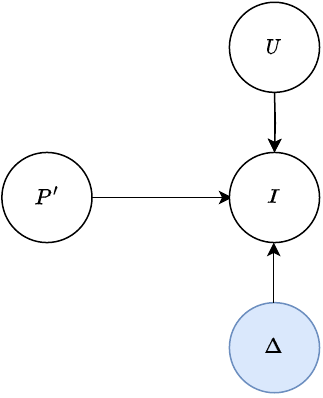}}  
    \hspace{+0.01in}
    \subfloat[Action \& Prediction]{ \includegraphics[width=0.15\textwidth]{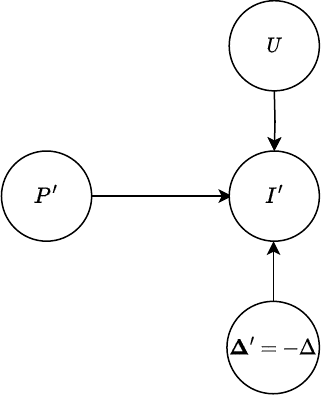}} 
    \vspace{-2.5mm}
    \caption{
        The proposed Doubly Abductive Counterfactual inference framework (DAC).
    }
\label{fig:dac}
\vspace{-6.5mm}
\end{figure}

\noindent\textbf{Our Solution}. 
To this end, we propose \textit{Doubly Abductive Counterfactual} inference framework (DAC). As illustrated in Figure~\ref{fig:dac}, following the formal three steps of counterfactual inference~\cite{pearl2016causal}: abduction, action, and prediction, we have:
\begin{itemize}
    \item \emph{Abduction-1}: $U = \arg\min_{U}\|G(P, U, \Delta=0)-I\|$.
    \item \emph{Abduction-2}: $\Delta = \arg\min_{\Delta}\|G(P', U, \Delta)-I\|$, where $\Delta$ transforms $P'$ back to $P$. 
    \item \emph{Action}: set $\Delta' = -\Delta$. 
    \item \emph{Prediction}: $I' = G(P', U, \Delta')$.
\end{itemize}
Our key insight stems from the newly introduced exogenous variable $\Delta$, which is the semantic change editing an imaginative $I'$ back to $I$. Although the overfitting of Abduction-2 also disables the natural language editability of $G$, it still enables the $\Delta$ editability. So, by reversing the change from $\Delta$ to $\Delta' = -\Delta$, we can use $\Delta'$ to edit $I$ back to $I'$. We detail the implementations of $U$ and $\Delta$ in Section~\ref{sec:method} and ablate them in Section~\ref{sec:ablation}. As shown in Figure~\ref{fig:visual_comparisions}, compared to existing methods, our DAC achieves a good trade-off between editability and fidelity, and thus we can support a wide spectrum of user editing intents including 1) addition, 2) removal, 3) manipulation, 4) replacement, 5) style transfer, and 6) face manipulation, which are extensively validated in both qualitative and quantitative evaluations in Section~\ref{sec:exp}. 
We summarize our contributions here:

\begin{figure*}[!htp]
    \centering
    \includegraphics[height=0.95\textheight]{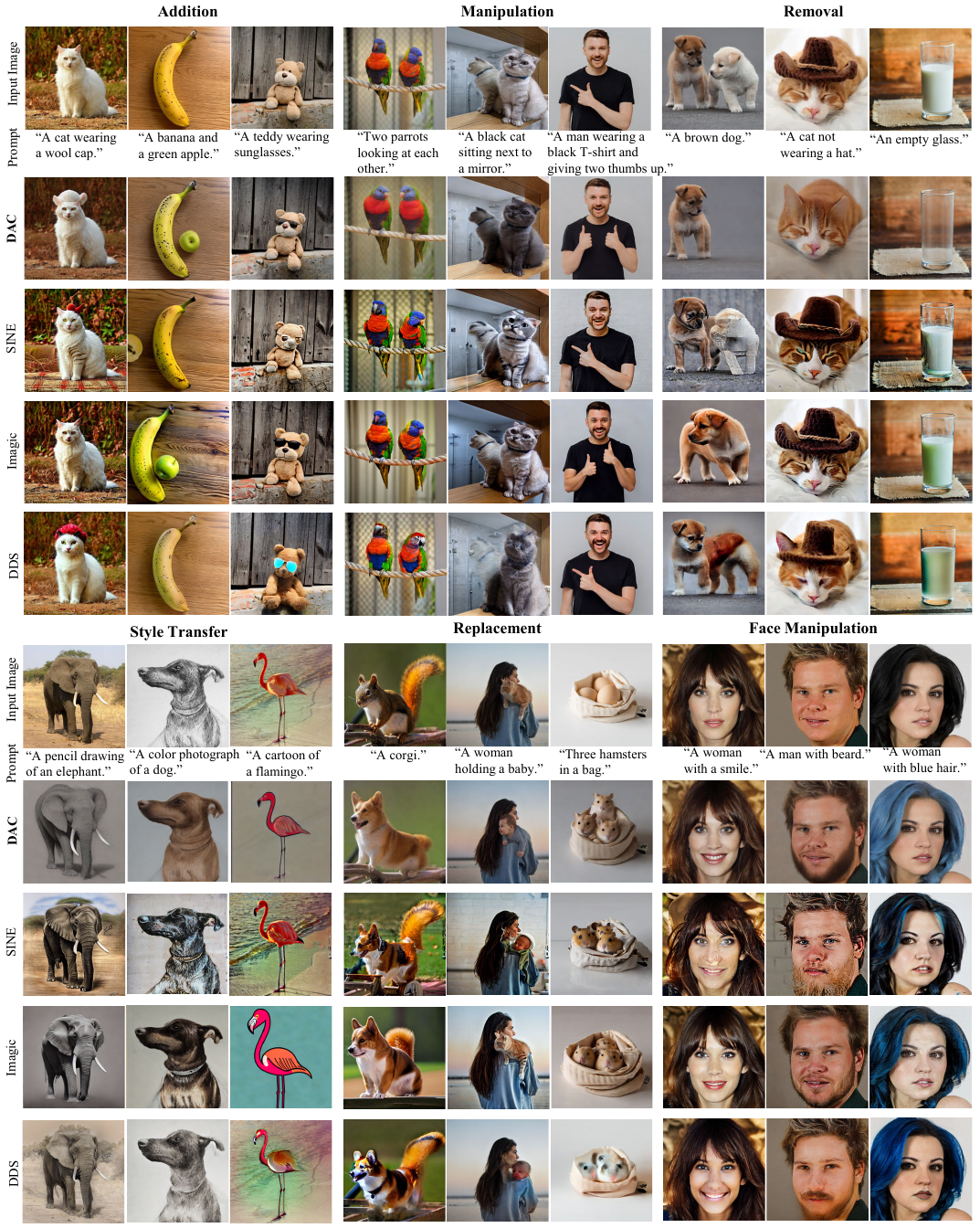}     
    \caption{Comparison of TBIE qualitative examples across the 6 editing types (only prompt $P'$ shown) between our DAC and three SOTAs with a similar design philosophy (Table~\ref{tab:editing_related_work}). For fairness, examples are chosen based on their best visual quality from various random seeds. See Section~\ref{sec:exp-1} for analysis and Appendix for the example selection details.}
\label{fig:visual_comparisions}
\end{figure*}

\begin{table*}[t]
\centering
\caption{Comparisons with existing methods.}
\vspace{-4mm}
\label{tab:as}
\resizebox{1.0\linewidth}{!}
{ 
\setlength{\tabcolsep}{2.2mm}{
\begin{tabular}{lcccc}
\toprule
Methods                               & $U$   & $\Delta$  & Method Description & Failure Analysis            \\ 
\midrule
P2P \cite{hertz2022prompt}            & \ding{55}     &\checkmark   &\multirow{7}{*}{\shortstack{$\Delta$ can be realized by adjusting \\ attention or feature maps}} & \multirow{7}{*}{\shortstack{Inversion methods are not accurate \\ for reconstruction w/o $U$}}  \\
TIME \cite{orgad2023editing}          & \ding{55}     &\checkmark   &   &  \\
PnP \cite{tumanyan2023plug}           & \ding{55}     &\checkmark   &   &  \\ 
MasaCtrl \cite{cao_2023_masactrl}     & \ding{55}     &\checkmark   &   &  \\
EDICT \cite{wallace2023edict}         & \ding{55}     &\checkmark   &   &  \\
AIDI \cite{pan2023effective}          & \ding{55}     &\checkmark   &   &  \\
CycleDiffusion \cite{wu2023latent}    & \ding{55}     &\checkmark   &   &  \\
\midrule
NTI \cite{mokady2023null}             & \checkmark    &\ding{55}   & \multirow{2}{*}{\shortstack{Modeling $U$ with textual inversion, \ie, fitting $I$ \\ with learnable text embeddings}} & \multirow{3}{*}{\shortstack{Editability is not enough \\for accurate editing w/o modeling $\Delta$}}   \\ 
PTI \cite{dong2023prompt}             & \checkmark  & \ding{55}   & & \\
\cdashline{1-4}
SINE \cite{zhang2023sine}             & \checkmark  & \ding{55}   &Modeling $U$ by textual inversion and fine-tuning SD & \\              
\midrule
DDS \cite{hertz2023delta}             & \checkmark  &\checkmark  &$U$ and $\Delta$ are learned together with the distillation loss & \multirow{2}{*}{\shortstack{$U$ and $\Delta$ are entangled, hard to find out the \\ best trade-off between the editability and fidelity}} \\
\cdashline{1-4}
Imagic \cite{kawar2023imagic}         & \checkmark  &\checkmark  &$U$ and $\Delta$ are learned by fine-tuning SD and textual inversion separately & \\
\midrule
DAC                                   & \checkmark &\checkmark   &Section~\ref{sec:method} &Section~\ref{sec:ablation} \\ 
\bottomrule
\end{tabular}}
}
\vspace{-4mm}
\label{tab:editing_related_work}
\end{table*}

\begin{itemize}
    \item We formulate text-based image editing (TBIE) into a counterfactual inference framework, which not only defines TBIE formally but also identifies its challenge: editability and fidelity trade-off.
    \item We propose the Doubly Abductive Counterfactual (DAC) to address the challenge.
    \item With extensive ablations and comparisons to previous methods, we demonstrate that DAC shows a considerable improvement in versatility and image quality.
\end{itemize}

\noindent \textbf{Notes. }In this paper, our purpose is to advocate that TBIE (or probably any visual editing) should be a counterfactual reasoning task, where the abduction is a necessary and crucial step. Unfortunately, we haven’t found a non-fine-tuning-based abductive learning method, and hence we conjecture that the absence of abduction is the key reason for the existing non-fine-tuning-based visual editing methods being fast yet not effective (e.g., SEED-LLaMA \cite{ge2023making}, Emu2 \cite{sun2023generative}, and InfEdit \cite{xu2023inversion}). Perhaps, only LLM can achieve both editing efficiency and effectiveness because LLM may perform counterfactual~\cite{tavares2021language}, but this requires unified vision-language tokens, which is in itself a challenging open problem.

\section{Related Work}
\label{sec:relatedwork}

\noindent\textbf{Text-to-Image Generation.}
The success of Imagen~\cite{saharia2022photorealistic} and DALL$\cdot$E~\cite{ramesh2022hierarchical} with diffusion models~\cite{ho2020denoising} opens a new era of open-domain text-to-image generation, being capable of generating diverse and high-quality images conditional on arbitrarily complex text descriptions. Thanks to the stable diffusion model~\cite{rombach2022high}, the text-to-image diffusion process could be conducted in a latent space of reduced dimensionality, bringing a significant speedup for training and inference. It is by far the most popular text-to-image model for open research, and thus we use a pre-trained one~\cite{rombach2022high} as our generative model $G$, although the proposed DAC framework is compatible with other generative models. 

\noindent\textbf{Text-based Image Editing.}
We summarize existing TBIE works in Table~\ref{tab:editing_related_work} from the perspective of counterfactual inference. We can see that they can be categorized into three groups based on whether $U$ and $\Delta$ are considered for both editability and fidelity. Note that we exclude other image editing methods like DreamBooth~\cite{ruiz2023dreambooth}, Cones2~\cite{liu2023cones}, and Textual inversion~\cite{gal2022image} that require multiple images for training, which are different from the TBIE settings covered in this paper. 

\noindent Group 1:  They directly operate the semantic change on the intermediate UNet attention maps during the generation process. The fidelity of the input image is achieved by DDIM inversion~\cite{tumanyan2023plug, cao_2023_masactrl} or other advanced inversion methods~\cite{wallace2023edict, pan2023effective, wu2023latent}, without explicitly modeling $U$.

\noindent Group 2: PTI~\cite{dong2023prompt}, NTI~\cite{mokady2023null}, and SINE~\cite{zhang2023sine} calculate $U$ by textual inversion or fine-tuning the stable diffusion model on the source image. Nevertheless, without $\Delta$, they cannot realize accurate editing, thus techniques like interpolation~\cite{dong2023prompt} are needed. 

\noindent Group 3: Imagic~\cite{kawar2023imagic} and DDS~\cite{hertz2023delta} learn $U$ and $\Delta$ together. However, the entanglement between $U$ and $\Delta$ makes it hard to find out the best trade-off between fidelity and editability.  

\noindent\textbf{Visual Counterfactuals.}
Counterfactual inference is the answer to a hindsight question like \textit{“When $Y=y$ and $X=x$, what
would have happened to $Y$ had $X$ been $x'$?”}. The general solution \cite{pearl2016causal} to the counterfactual inference is to abduct the exogenous variables with the known fact ($Y=y, X=x$) and then reset our choice ($X=x'$) and obtain the new prediction ($Y=\;?$). Counterfactual inference has a wide application in computer vision such as visual explanations \cite{goyal2019counterfactual}, data augmentations \cite{kaushik2019learning}, robustness \cite{simon1954spurious, balashankar2021can, tang2020long}, fairness \cite{kusner2017counterfactual, zhang2018fairness}, and VQA \cite{niu2021counterfactual}.

\section{Method}
\label{sec:method}

Recall in Section~\ref{sec:introduction} that our proposed Doubly Abductive Counterfactual inference framework (DAC) is to address the non-editability issue caused by the overfitted abduction of $U$ that was originally introduced for the purpose of keeping minimal visual change. This issue is elegantly resolved by introducing another abduction of a semantic change variable $\Delta$. In this section, we will detail the implementation of every step in DAC as illustrated in Figure~\ref{fig:dac}. 

\subsection{Abduction-1}
We introduce the implementation of the abduction loss $\|G(P, U, \Delta = 0) - I\|$. This step is identical to the conventional abduction of $U$ in Figure~\ref{fig:counterfactual_inference}, as we set $\Delta = 0$ in Figure~\ref{fig:dac} (a). In particular, we use Stable Diffusion~\cite{rombach2022high} to implement $G$ due to it being open-source and for a fair comparison with other methods. As $\|G(P, U, \Delta = 0) - I\|$ is essentially a reconstruction loss, we abduct $U$ by solving the following Gaussian noise regression as in training the reversed diffusion steps:
\begin{equation}
\arg\min\limits_{U}\mathbb{E}_{(t,\epsilon)} ||  \epsilon - \Theta_{(U,\Delta=0)}(x_{t},t,P) ||_2^2,  \label{eq:abduction_1}
\end{equation}
where $\epsilon\in\mathcal{N}(\mathbf{0},\mathbf{I})$, $t\in[0,T]$ is a sampled time step ($T$ is the maximum), $\Theta_{(U,\Delta=0)}$ is the pre-trained noise prediction UNet with trainable $U$ and all other parameters frozen, conditionally on language tokens of $P$ encoded by a frozen CLIP~\cite{radford2021learning} text encoder\footnote[2]{As $\Delta$ is also a LoRA (Section~\ref{sec:abduction_2}), $\Delta = 0$ corresponds to the original, unmodified encoder.}, $x_{t} = \sqrt{\alpha_{t}} x_{0} + \sqrt{1-\alpha_{t}} \epsilon$ is the noisy input at $t$, in particular, $x_0 = I$, and $\alpha_{t}$ is related to a fixed variance schedule \cite{ho2020denoising, song2020denoising}.

\begin{figure}[t]
    \centering
\includegraphics[width=0.46\textwidth]{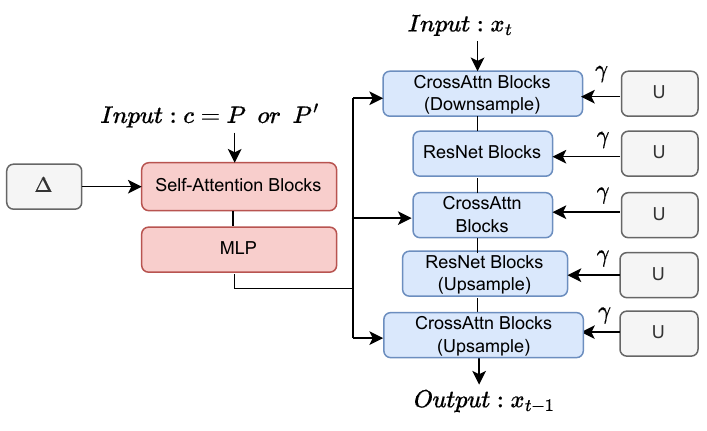} 
    \vspace{-0.2in}
    \caption{Parameterizations of $U$ and $\Delta$ by using LoRA (grey) for UNet (blue) and text encoder (red) in pre-trained Stable Diffusion~\cite{rombach2022high}: $\Theta_{(U, \Delta)}(x_t, t, c)$. Except for LoRA, all the other parameters are frozen.}
    \vspace{-5mm}
\label{fig:lora}
\end{figure}

We parameterize $U$ as the UNet LoRA~\cite{hu2021lora} in $\Theta_{(U,\Delta)}$. As shown in Figure~\ref{fig:lora}, the LoRA structure is built on all of the attention layers, convolutional layers, and feed-forward (FFN) layers. This is because we observe the underfitting issue if we only apply LoRA on the attention layers, \ie,  $I$ cannot be well-reconstructed using $P$ and $U$ (See ablation in Appendix).

Without loss of generality, we only detail the implementation of a linear layer with a LoRA structure. Denote $z \in \mathbb{R}^{d}$ as the intermediate feature, $W \in \mathbb{R}^{d \times d}$ as the parameter of the linear layer,  then the output $z'$ after LoRA becomes:
\begin{equation}
    z' = (W + U_{A} \cdot U_{B}) \cdot z,
\end{equation}
where $U_{A} \in \mathbb{R}^{d \times r}$ and $U_{B} \in \mathbb{R}^{r \times d}$ are low rank matrices with $r<d$.

\subsection{Abduction-2}
\label{sec:abduction_2}
We introduce the implementation of the second abduction loss $\|G(P', U, \Delta)-I\|$ with the above abducted $U$ (Figure~\ref{fig:dac} (b)). Similar to Eq.~\eqref{eq:abduction_1}, we minimize:
\begin{equation}
\arg\min\limits_{\Delta}\mathbb{E}_{(t,\epsilon)} ||  \epsilon - \Theta_{(U, \Delta)}(x_{t},t,P') ||^2,  \label{eq:abduction_2}
\end{equation}
where we parameterize $\Delta$ as the CLIP text encoder LoRA, and $U$ calculated in Abduction-1 is frozen.

As shown in Figure~\ref{fig:lora}, the LoRA structure is only built on the attention layers of the CLIP text encoder.
The self-attention layer language feature $y'$ in the CLIP text encoder is re-encoded from the original $y$ through the LoRA:
\begin{equation}
    y' = (W + \Delta_{A} \cdot \Delta_{B}) \cdot y,
    \label{eq:text_lora}
\end{equation}
where $\Delta_{A} \in \mathbb{R}^{d \times r}$ and $\Delta_{B} \in \mathbb{R}^{r \times d}$ are low rank matrixes, $r<<d$. By solving Eq.~\eqref{eq:abduction_2}, $\Delta$ encodes the visual transition controlled by $P'$ to $P$. We highlight that $\Delta$ cannot be parameterized by textual inversion~\cite{mokady2023null}, as it does not support semantic inversion as introduced later in Section~\ref{sec:method_prediction}.

If $U$ is overfitted in Abduction-1, \eg, $U$ memorizes everything about $I$, the Abduction-2 for $\Delta$ might be as trivial as $\Delta = 0$. Inspired by the findings in diffusion models where a larger time step corresponds to better editability while lower fidelity~\cite{wang2023not},  we design an annealing strategy on $U$ in solving Eq.~\eqref{eq:abduction_2} at different time steps:
\begin{eqnarray}
    z' &=& (W + \gamma U_{A} \cdot U_{B}) \cdot z, \\
    \gamma &=& \frac{1-\eta}{T^{2}} (t-T)^{2} + \eta,\label{eq:8}
\end{eqnarray}
where $\eta \in \mathbb{R}$ is a small constant value. In general, $\eta$ is a hyper-parameter dependent on both $I$ and $(P, P')$; fortunately, it is easy to choose a good one as shown in Figure~\ref{fig:exp_annealing}.

\subsection{Action \& Prediction}
\label{sec:method_prediction}
We introduce the implementation of action \& prediction procedures $I' = G(P', U, \Delta')$ in Figure~\ref{fig:dac} (c). 
First, we take the action $\Delta'=-\Delta$ to revert the visual transition from $P$ back to $P'$ to generate $I'$. Thus, the text LoRA in Eq.~\eqref{eq:text_lora} becomes:
\begin{equation}
    y' = (W - \Delta_{A} \cdot \Delta_{B}) \cdot y.
    \label{eq:prediction_lora}
\end{equation}
Then, with a sampled $x_{T} \in \mathcal{N}(0,\mathbf{I})$, the DDIM sampling~\cite{song2020denoising} is used to generate the edited image $I'$ with the following iterative update from $t = T$ to $t = 0$:
\begin{eqnarray}
    x_{t-1} &= \sqrt{\alpha_{t-1}} \left( \cfrac{x_{t} - \sqrt{1-\alpha_{t}}\Theta_{(U, \Delta')}(x_{t}, t, P') }{\sqrt{\alpha_{t}}} \right) +  \nonumber\\ 
    &\sqrt{1-\alpha_{t-1}} \Theta_{(U, \Delta')}(x_{t},t,P'),
\end{eqnarray}
where we obtain $I'=x_{0}$. Interestingly, as shown in Figure~\ref{fig:delta_weight}, we use a weight $\beta\in[-1,1]$ to tune $\beta\Delta_{A} \cdot \Delta_{B}$ in Eq.~\eqref{eq:prediction_lora}  to manifest the inversion ability of $\Delta$, where $\beta = -1$ means reconstruction of the source image as in Eq.~\eqref{eq:text_lora} and $\beta > -1$ means that we start to shift the semantic change from the source image.

\section{Experiment}
\label{sec:exp}
We followed prior works~\cite{tumanyan2023plug, cao_2023_masactrl, wallace2023edict, pan2023effective, wu2023latent, dong2023prompt, kawar2023imagic, hertz2023delta} to use Stable Diffusion as our generator~\cite{rombach2022high}. For fair comparisons, we integrated SD checkpoint V2.1-Base with the official source codes of the comparing methods: SINE~\cite{zhang2023sine}, DDS~\cite{hertz2023delta}, and Imagic~\cite{kawar2023imagic} in the Diffusers codebase~\cite{von-platen-etal-2022-diffusers} and we used the same default hyper-parameters of the SDV2.1-Base. In particular, during the optimization of $U$ and $\Delta$ in Abduction-1 and Abduction-2, we set the rank of the LoRA to 4 for $\Delta$ and 512 for $U$, the learning rate to 1e-4. Optimization iterations were 1,000 in both Abduction-1 and Abduction-2. $\eta \in [0.4,0.8]$ is applied to the annealing strategy. 
For the action and prediction steps, we adopted 30 steps for DDIM sampling at the inference time of the stable diffusion. We used an NVIDIA A100 GPU for editing. 

\noindent \textbf{Computation Analysis. }In general, it took 120, 0.33, 12, and 15 minutes to edit a single image by using SINE, DDS, Imagic, and our DAC. Our method consumes 15 minutes, including 6 and 9 minutes for the first and second abduction, and 4-second 30-step DDIM sampling. The time-saving characteristic of DDS lies in minimal trainable parameters (latent format of an image in DDS compared with UNet LoRA or CLIP text encoder LoRA in DAC's abduction) and minimal optimization iterations (200 iterations in DDS compared with 1,000 iterations in DAC).

\begin{figure}[t]
    \centering
\includegraphics[width=0.48\textwidth]{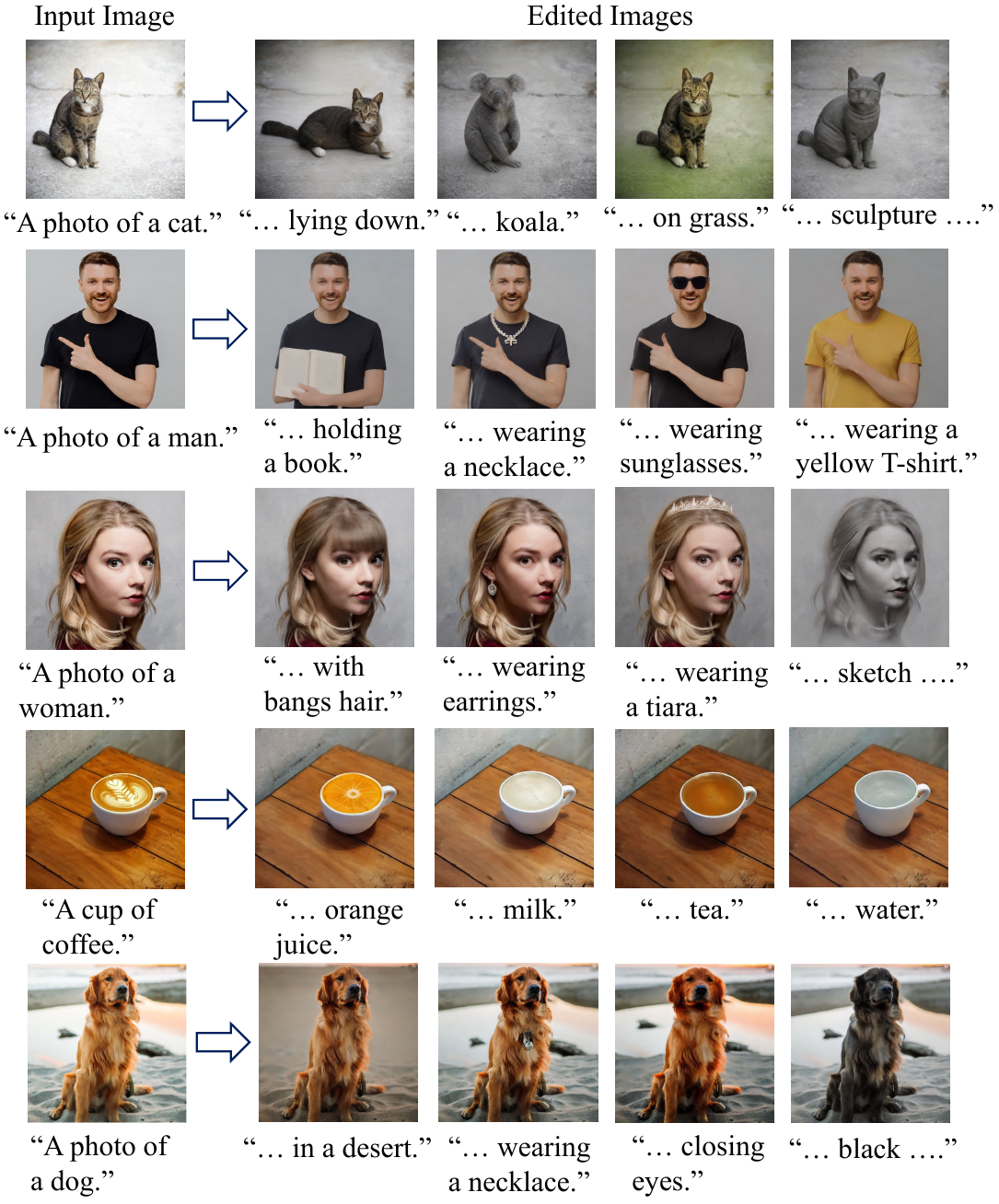} 
    \vspace{-6mm}
    \caption{Qualitative examples of DAC with different prompts editing on the same source image.}
    \vspace{-6mm}
\label{fig:multi-edit}
\end{figure}

\subsection{Qualitative Evaluation}
\label{sec:exp-1}
We demonstrate the advantages of the proposed DAC method with two kinds of qualitative evaluations:
1) evaluation of our method with multiple prompts on the same source image, and 2) evaluation of our method on the 6-type editing operations. For each editing, we randomly generated 8 edited images given a source image and an editing prompt, and chose the one with the best quality as our final edited image.
Note that such a process is also adopted for other comparison methods. 
Following previous works \cite{kawar2023imagic, zhang2023sine}, we collected most images from a wide range of domains, \ie, free-to-use high-resolution images from Unsplash (https://unsplash.com/).

\noindent{\bf Editing with Multiple Prompts.} As shown in Figure~\ref{fig:multi-edit}, we generate the edited images with a source image and multiple editing prompts. With a photo of a man, we enable him to hold a book, wear a necklace, wear sunglasses, or change the black shirt to a yellow one, while keeping a good fidelity of the source image. It also shows that our DAC enjoys impressive editing ability when applied to various images with different language guidance, manifesting the good versatility of our method.

\noindent{\bf Wide Spectrum of Editing.} We demonstrate that our DAC supports a wide spectrum of editing operations including 1) addition, 2) removal 3) manipulation, 4) replacement, 5) style transfer, and 6) face manipulation. Our results are summarized in Figure~\ref{fig:visual_comparisions} and more results are in Appendix. For one of the 6 editing types, we provide three image-prompt examples. Take an example for manipulation, we make two parrots look at each other, change the white cat with its mirror to a black one, and let a man give two thumbs up. After the editing, the images not only resemble the source image to a high degree but also are coherent with the text prompt, demonstrating that the DAC method achieves a great trade-off between fidelity and editability.

\begin{figure}[t]
    \centering
    \includegraphics[width=0.42\textwidth]{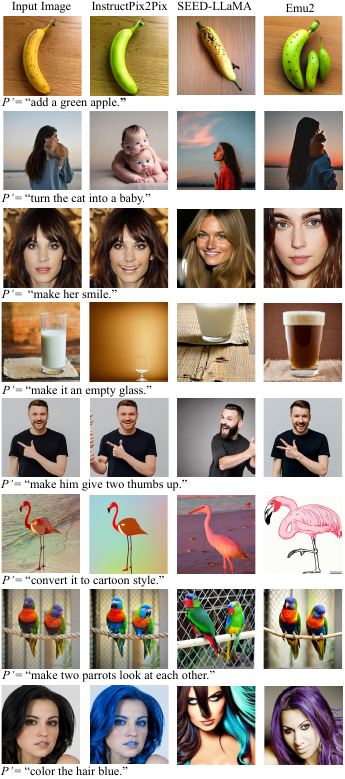} 
    \vspace{-1mm}
    \caption{Qualitative examples of large-scale training methods.}
\label{fig:others}
\vspace{-6mm}
\end{figure}

\begin{figure}[t]
    \centering
 \includegraphics[width=0.48\textwidth]{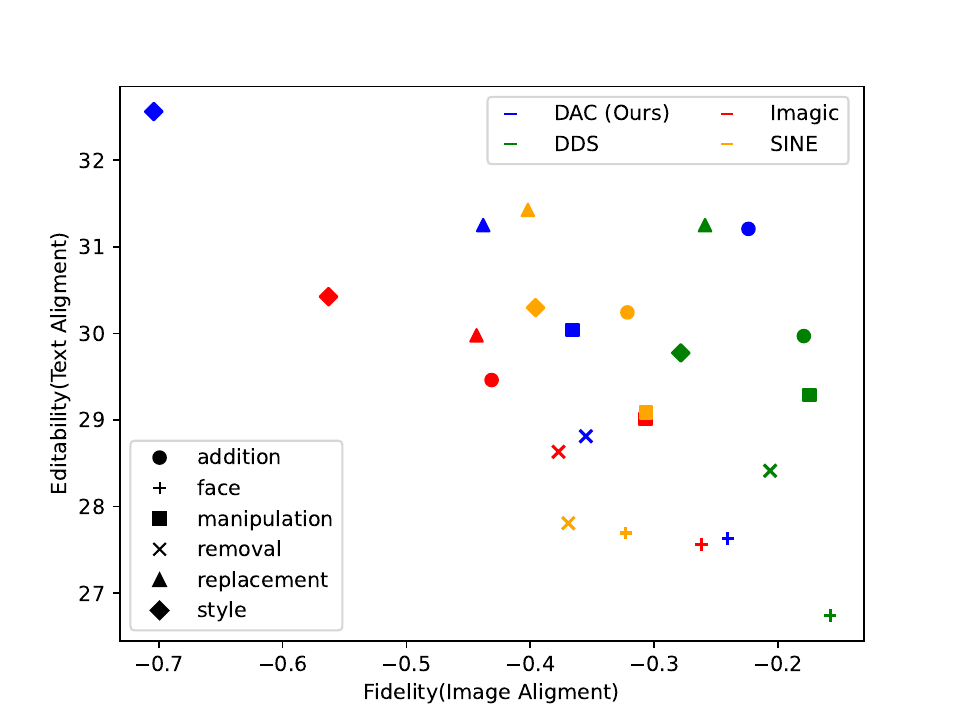} 
    \vspace{-6mm}
    \caption{Image Alignment: minus LPIPS. Text Alignment: CLIP-score. Both values are the larger the better.}
    \vspace{-4mm}
\label{fig:compare_quan}
\vspace{-2mm}
\end{figure}

\noindent{\bf Comparisons with Competitive Methods.}
We compare DAC with leading works on the TBIE task including Imagic~\cite{kawar2023imagic}, SINE~\cite{zhang2023sine}, and DDS~\cite{hertz2023delta}.
And they all belong to single-image fine-tuning methods for a fair comparison.
To have a more comprehensive understanding of the superiority of the DAC method, we compare it with the three methods in the 6 kinds of editing operations in  Figure~\ref{fig:visual_comparisions}.
Compared with previous methods, the DAC method enjoys the following merits.
First, the generated images by the DAC method are more consistent with the textual prompts. 
With prompts such as ``remove the milk in the glass", and ``let two parrots look at each other", our method successfully makes it while it is hard for previous methods.
Second, the DAC method can keep better fidelity to the source image.
With prompts like ``replace the squirrel with a corgi" and ``remove the white dog", the edited images by the DAC resemble the input images to a much higher degree than previous methods.
All of these samples in Figure~\ref{fig:visual_comparisions} indicate that the DAC method does a better trade-off between fidelity and editability, achieving state-of-the-art performance on the TBIE task.

In addition to single-image fine-tuning methods, there are works that conduct large-scale training and don't require any test-time fine-tuning, e.g., InstructPix2Pix \cite{brooks2023instructpix2pix}, SEED-LLaMA \cite{ge2023making}, and Emu2 \cite{sun2023generative}. We have shown that the ``fine-tuning'' is the essential ``abduction'' for fidelity. However, these methods only have inference-time editing---only ``action'' and ``prediction'', thus they cannot guarantee fidelity in theory (as depicted in Figure \ref{fig:others}).

\subsection{Quantitative Evaluation}
\noindent\textbf{CLIP-score~\cite{radford2021learning} and LPIPS~\cite{zhang2018unreasonable}.}
The experimental settings were set as follows.
\begin{itemize}
    \item Different editing operations need different trade-offs between fidelity and editability. For example, style transfer requires lower image alignment compared to object manipulation. Thus, the evaluations of six kinds of editing are conducted individually.
    \item We applied 9 different prompt-image pairs for each kind of editing.
    \item We calculated LPIPS for the image alignment and CLIP-score for text alignment.
\end{itemize}

We summarize the results in Figure~\ref{fig:compare_quan}.
The proposed DAC method shows better performance in text alignment scores for editing like object removal, object manipulation, object addition, and face manipulation. We achieved similar results with the DDS~ \cite{hertz2023delta} in object replacement. For the style transfer, DAC achieves the best text alignment scores. The LPIPS score measures the image alignment degree between the source image and the edited image. 
However, we argue that LPIPS fails to reflect the fidelity. For example in Figure~\ref{fig:visual_comparisions}, ``remove the hat of the cat". Our DAC successfully removes the hat and achieves a better CLIP-score. DDS and SINE methods cannot remove the hat and thus have a lower CLIP-score. But DDS and SINE achieve a much higher LPIPS score because they make no changes at all to the source image. Therefore, we have to conduct a user study for a more accurate assessment.

\begin{wrapfigure}{r}{0.17\textwidth}
    \centering
    \includegraphics[width=.17\textwidth]{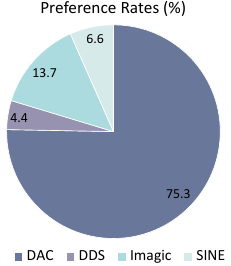}
    \vspace{-7mm}
    \caption{User study statistics.}
    \label{fig:user}
    \vspace{-5mm}
\end{wrapfigure}

\noindent{\bf User Study.}
In this section, we quantitatively evaluate our DAC with an extensive human perceptual evaluation study. 
First, we collected a diverse set of image-prompt pairs, covering all the ``addition", ``manipulation", ``removal", ``style transfer", ``replacement", and ``face manipulation" editing types. It consists of 54 input images and their corresponding target prompts. 
110 AMT participants were given a source image, a target prompt, and 4 edited images by DAC, DDS, SINE, and Imagic, which were randomly shown. The participants are required to choose the best-edited image. In total, we recalled 5,940 answers. The result is summarized in Figure~\ref{fig:user} and it shows that 75.3\% evaluators preferred our DAC. The user interface is detailed in Appendix.

\begin{figure}[t]
    \centering    \includegraphics[width=0.48\textwidth]{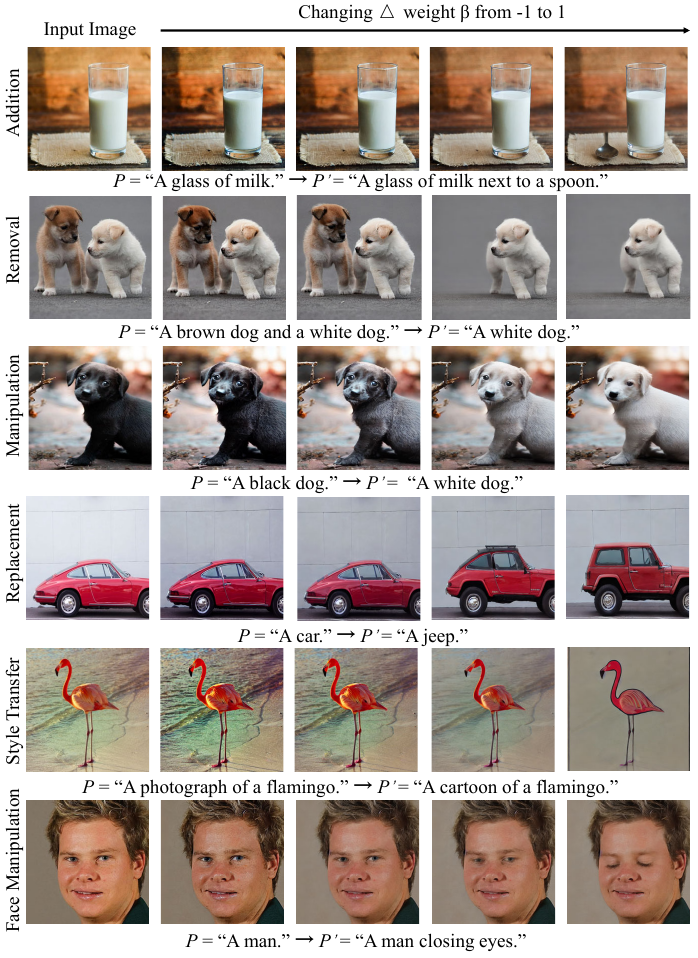} 
    \vspace{-8mm}
    \caption{Illustrations of ablating the weight $\beta$ for $\beta\Delta_A\cdot\Delta_B$ in Eq.~\eqref{eq:prediction_lora}.}
    \vspace{-6mm}
\label{fig:delta_weight}
\end{figure}

\begin{figure}[t]
    \centering
\includegraphics[width=0.48\textwidth]{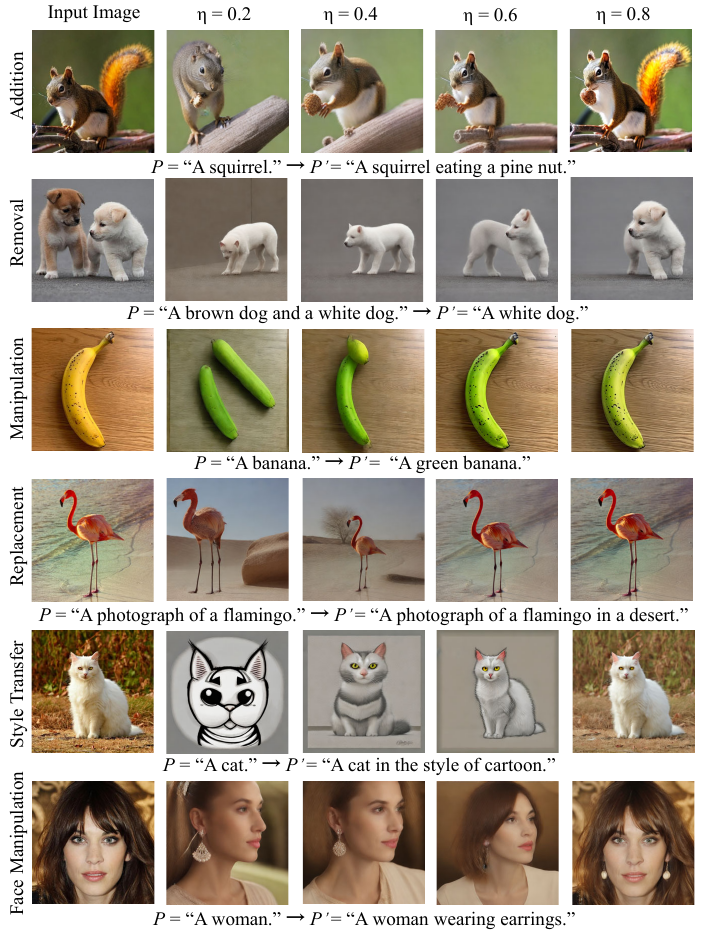} 
    \vspace{-8mm}
    \caption{Illustrations of ablating the annealing hyper-parameter $\eta$ in Eq.~\eqref{eq:8}.}
    \vspace{-4mm}
\label{fig:exp_annealing}
\end{figure}

\subsection{Ablation Analysis}
\label{sec:ablation}
\noindent{\bf Training Iterations and Editability.}
We examined the relationship between training iterations of $\arg\min_{U} \|G(P, U)- I\|$ and editability by applying six different types of editing operations.
As shown in Figure~\ref{fig:editability_interations}, with the dog image and the prompt ``A dog. $\to$ A jumping dog", we can get a jumping dog in the edited image using 250 and 500 training iterations.  However, the images are with low fidelity. Training $U$ in 1000 iterations, the generative model fails to make the dog jump and the edited image looks the same as the source one, implying good fidelity but poor editability. 
This study indicates that with the increase of training iterations $\arg\min_{U} \|G(P, U)- I\|$, the editability decreases while the fidelity increases, which means a good $U$ is needed for the best trade-off between fidelity and editability. This is exactly the challenge of the TBIE task.

\noindent{\bf Ablation on $\Delta$ Subtraction.}
In the procedure of action \& prediction $I' = G(P', U, \Delta')$, the $\Delta$ is reversed to $\Delta'=-\Delta$. We use $\Delta'$ to edit $I$ back to $I'$. Nevertheless, considering $\Delta'=-\beta \Delta$, there could be different $\beta$ values. We examined the effects of $\beta$ values on the edited image.
As shown in Figure~\ref{fig:delta_weight}, with the black dog image and the prompt ``A black dog $\to$ A white dog", increasing $\beta$ from -1 to 1, the black dog changes to a gray one first and then a white one.
Similarly, A car can be smoothly edited into a Jeep as $\beta$ varies from -1 to 1.
From the examples in Figure~\ref{fig:delta_weight}, the learned $\Delta$ can be considered as the direction vector of our desired semantic change. Different $\beta$ values imply different strengths to apply the semantic change. However, for rigid manipulations like Addition and Removal, $\beta$ does not show a gradual transition, which is reasonable as it is hard to quantify the existence level of an object.

\noindent{\bf Ablation on Annealing Strategy.}
We ablated the annealing strategy in the Abduction-2 step.
As shown in Figure~\ref{fig:exp_annealing}, we observe that $\eta \in [0.4,0.8]$ is a reasonable interval for successful editing. A larger time step in the stable diffusion model corresponds to better editability while lower fidelity. 
The smaller $\eta$ indicates that we leverage more priors of the pre-trained weights at large time steps, thus increasing the editability while decreasing the fidelity. 
This is consistent with the phenomenon in Figure~\ref{fig:exp_annealing}: as $\eta$ increases from 0.2 to 0.8, the edited images show better fidelity to the source images although the editability decreases. With $\eta \in [0.4, 0.8]$, we achieve a good trade-off between fidelity and editability.

\begin{figure}[t]
    \centering
\includegraphics[width=0.42\textwidth]{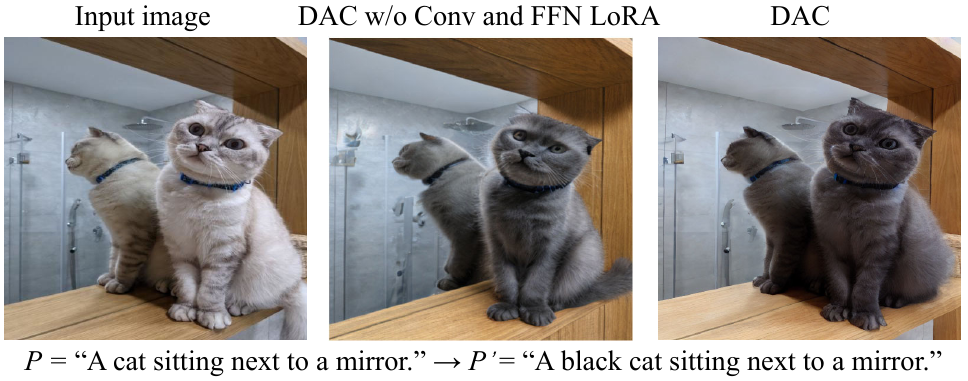} 
    \vspace{-2mm}
    \caption{Ablation on UNet LoRA.}
\label{fig:ablation_conv}
\vspace{-6mm}
\end{figure}

\noindent{\bf Ablation on UNet LoRA.}
For $U$, we added the LoRA structure on all of the attention layers, convolutional layers, and FFN layers to guarantee the fidelity of the source image. As shown in Figure~\ref{fig:ablation_conv}, if we only used attention layers LoRA, the editing would blur the background details; after adding LoRA to convolutional layers and FFN layers, we can retain the details successfully. 

\begin{figure}[t]
    \centering
\includegraphics[width=0.42\textwidth]{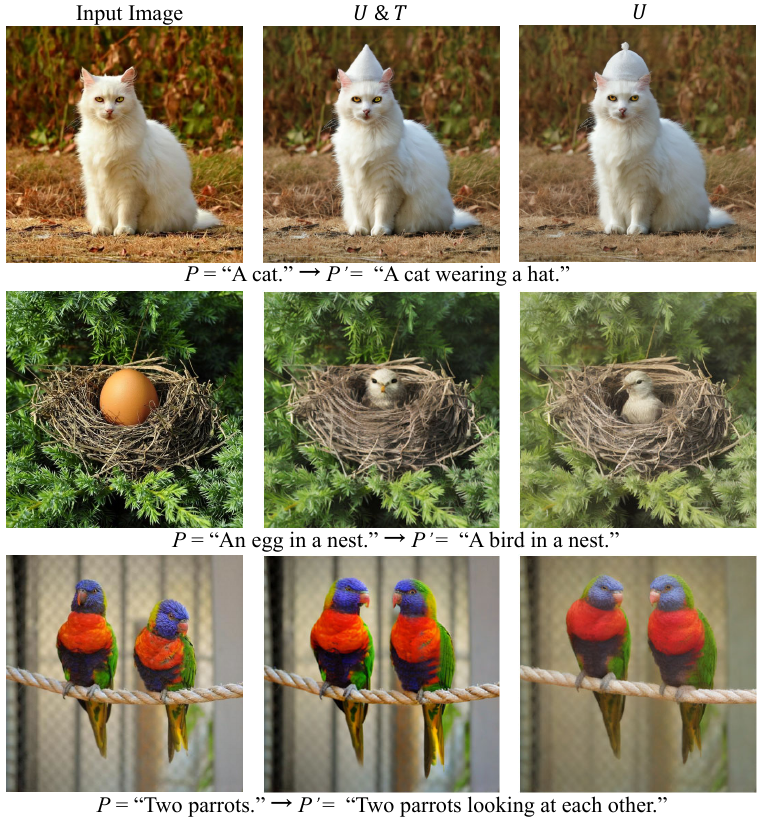} 
    \vspace{-0.1in}
    \caption{Ablation on Abduction-1.}
    \vspace{-3mm}
\label{fig:ablation_ab1}
\end{figure}

\noindent{\bf Ablation on Abduction-1.}
In the Abduction-1, we abduct $U$ to encode the content of image $I$, thus guaranteeing a good fidelity. However, since images contain various contents, the $U$ abducted from the same settings (e.g., training iterations) may not be able to achieve an overfit encoding for some complex images. Then the remaining information will be abducted in $\Delta$. When we take the action $\Delta'=-\Delta$ and implement prediction, such information will be subtracted, leading to information loss in $I'$ (as shown in the third column in Figure~\ref{fig:ablation_ab1}). To make a complement for such information, we could introduce another exogenous variable $T$ parameterized as the CLIP text encoder LoRA, which satisfies $\arg\min_{T}\|G(P, U, T, \Delta=0)-I\|$. Finally, the prediction becomes $I' = G(P', U, T, \Delta')$ (the second column in Figure~\ref{fig:ablation_ab1}). It could be seen that the incorporation of $T$ in the Abduction-1 achieves a better fidelity than the abduction of $U$ only. Moreover, conducting iterative abduction on $U$ and $T$ more times could further improve fidelity. Considering that the abduction of $U$ is enough for most cases and the computation cost produced by the abduction of $T$, we only adopt $U$ in our experiments.

\begin{figure}[t]
    \centering
    \includegraphics[width=0.42\textwidth]{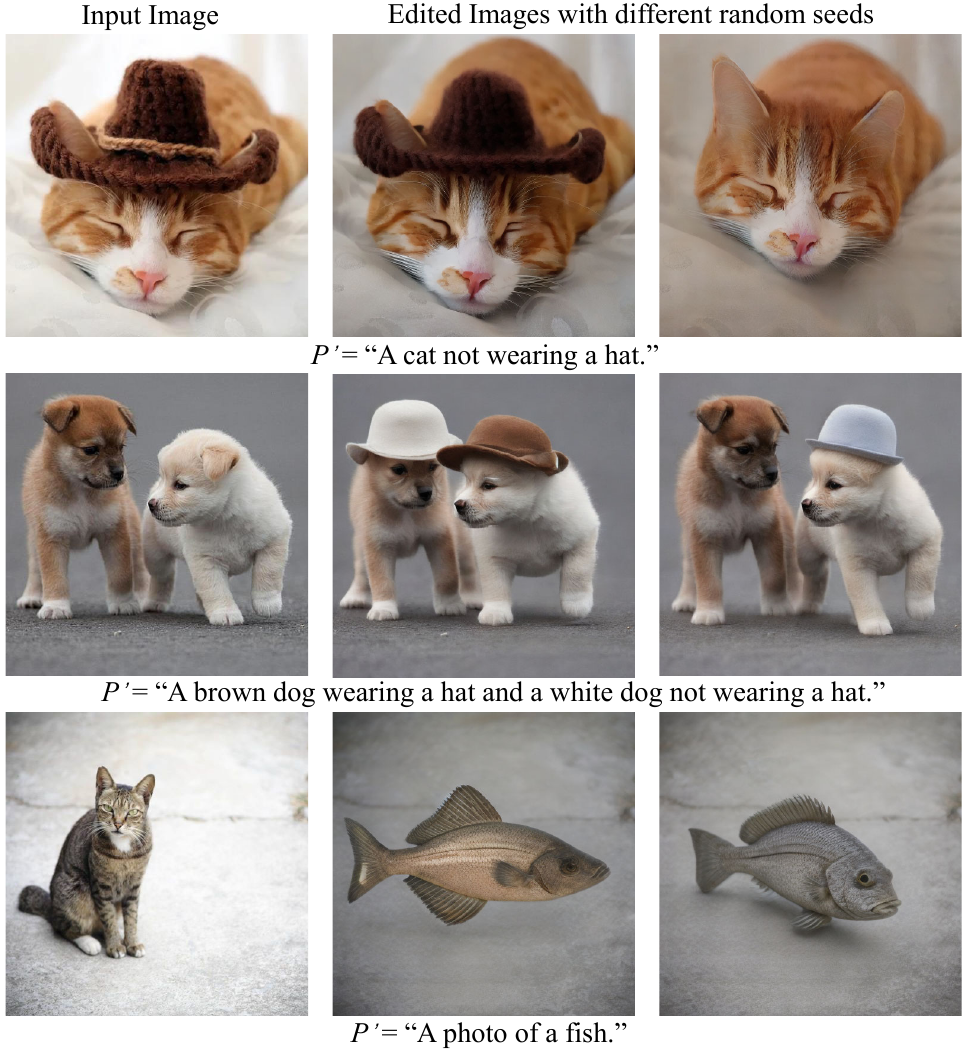} 
    \vspace{-0.1in}
    \caption{Failure cases due to the issues in stable diffusion.}
    \vspace{-4mm}
\label{fig:failure}
\vspace{-2.8mm}
\end{figure}

\noindent\textbf{Failure Case Study.}
We observed three kinds of failures caused by stable diffusion: 1) sensitivity to random seeds, 2) the incapability of comprehending referring expressions, and a more subtle case 3) the lack of common sense. As shown in Figure~\ref{fig:failure}, random seed impacts the success rate of generation; the second failure is due to the fact that stable diffusion cannot always generate images according to the prompt with referring expressions like ``a white dog next to a brown dog and the brown dog is wearing a hat''; if we change the object from cat to fish, we should also change the background from land to water due to the common sense ``fish lives in water''. To fundamentally resolve such failures, maybe we need to improve stable diffusion to endow such capabilities. We leave it to future work. 
\section{Conclusions}
We proposed to formulate the task of TBIE using a theoretical framework: counterfactual inference, which clearly explains why the challenge is the trade-off between editability and fidelity: the overfitted abduction of the source image parameterization, which is a single-image reconstruction fine-tuning. To this end, we propose Doubly Abductive Counterfactual (DAC). The key idea is that, since we cannot avoid the overfitting of the above abduction, we use another overfitted abduction, which encodes the semantic change of the editing, to reverse the lost editability caused by the first one. We conducted extensive qualitative and quantitative evaluations on DAC and other competitive methods. Our future work is two-fold. First, we will upgrade DAC to support visual example-based editing~\cite{ruiz2023dreambooth,liu2023cones}. Second, we will use Fast Diffusion Model \cite{wu2023fast} and Consistency Models~\cite{song2023consistency} such as Latent Consistency LoRA~\cite{luo2023latent} to speed up the fine-tuning and inference in editing.

\noindent \textbf{Acknowledgements. }This work was supported by NSFC project (No. 62232006), in part by Shanghai Science and Technology Program (No. 21JC1400600), and by National Research Foundation, Singapore under its AI Singapore Programme (AISG Award No: AISG2-RP-2021-022).
{
    \small

    \bibliographystyle{ieeenat_fullname}
}

\clearpage
\appendix

\noindent The \textbf{Appendix} is organized as follows:
\begin{itemize}[leftmargin=*]
    \item \textbf{Section~\ref{a}:} provides more details on sample selection process and shows more experimental results, e.g., quantitative and qualitative results on InsturctPix2Pix dataset.
    \item \textbf{Section~\ref{b}:} gives more ablation analysis.
    \item \textbf{Section~\ref{c}:} provides more details on user study.
    \item \textbf{Section~\ref{d}:} provides analysis on the limitation.
\end{itemize}

\section{Additional Results}
\label{a}
\noindent{\bf Selection Details.} 
In the main paper and Appendix, we use the following sample selection process for all comparative methods.
(1) The hyperparameter $\eta \in \{0.4, 0.6, 0.8\}$ is adopted in DAC. (2) The hyperparameter for text embedding interpolation in Imagic~\cite{kawar2023imagic} is in $\{0.9, 1.2, 1.4\}$. 
(3) DDS~\cite{hertz2023delta} sets different numbers of the classifier free guidance scale as 3, 5, and 7.5 respectively. 
In addition to different values of hyperparameters, for each editing, we randomly generated 8 edited images given a source image and an editing prompt and chose the one with the best quality as the final edited image.

\subsection{Quantitative Results}
To further evaluate the effectiveness of DAC, we leverage a random subset of 200 paired prompts and images in the InsturctPix2Pix dataset~\cite{brooks2023instructpix2pix}. 
Considering that SINE~\cite{zhang2023sine} requires a huge time cost for a single image editing (\ie, 2 hours), we exclude it from this comparison. The results are listed in Table \ref{tab:supp_compare}. DDS~\cite{hertz2023delta} obtains the best image alignment with source images (\ie, the lowest LPIPS score) while the worst text alignment with prompts (\ie, the lowest CLIP-score). This is because DDS~\cite{hertz2023delta} mostly makes no change to source images, thus failing to achieve effective editing.
Compared with Imagic~\cite{kawar2023imagic}, our DAC archives a lower LPIPS score and a higher CLIP-score, which demonstrates higher fidelity to source images and better editability. Therefore, DAC fulfills a better trade-off between fidelity and editability for text-based image editing.

\begin{table}[b]
\centering
\vspace{-0.5cm}
\caption{Quantitative comparisons on InsturctPix2Pix dataset~\cite{brooks2023instructpix2pix}.}
\vspace{-0.25cm}
\label{tab:as}
\begin{tabular}{lcccc}
\toprule
Methods              & \textbf{DAC}   & Imagic~\cite{kawar2023imagic}  & DDS~\cite{hertz2023delta}            \\ 
\midrule
LPIPS $\downarrow$        & 0.40 & 0.43 & 0.24         \\ 
CLIP-score $\uparrow$     &32.3 & 31.4  & 30.8        \\
\bottomrule
\end{tabular}
\label{tab:supp_compare}
\end{table}

\begin{figure}[t]
    \centering
    \includegraphics[width=0.48\textwidth]{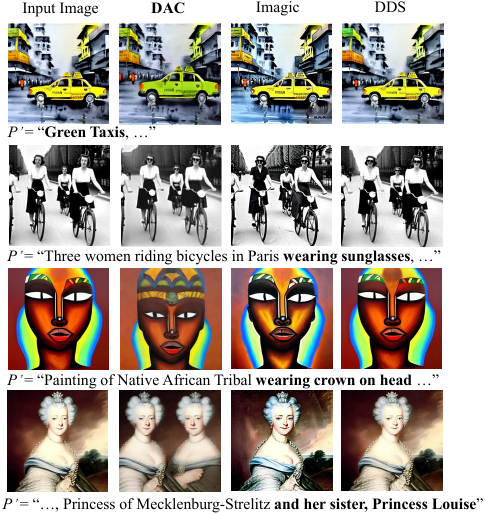} 
    \vspace{-6.5mm}
    \caption{Visual comparisons on InsturctPix2Pix dataset~\cite{brooks2023instructpix2pix}.}
\label{fig:supp_comp}
\vspace{-7mm}
\end{figure}

\begin{figure*}[!htp]
    \centering
    \includegraphics[height=0.95\textheight]{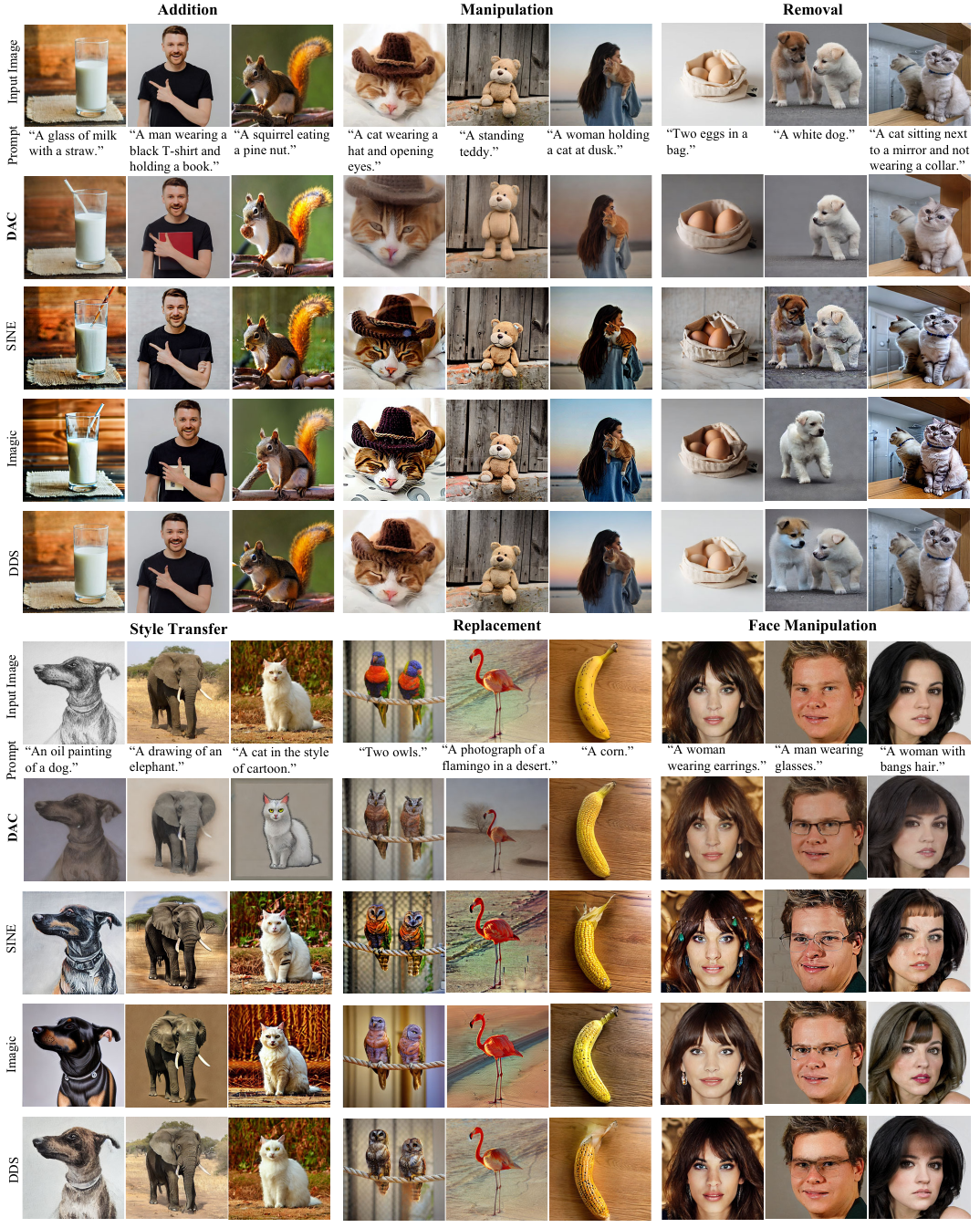}     
    \caption{Comparison of TBIE qualitative examples across the 6 editing types (only prompt $P'$ shown) between our DAC and three SOTAs.}
\label{fig:comparisions1}
\end{figure*}

\begin{figure*}[!htp]
    \centering
    \includegraphics[height=0.95\textheight]{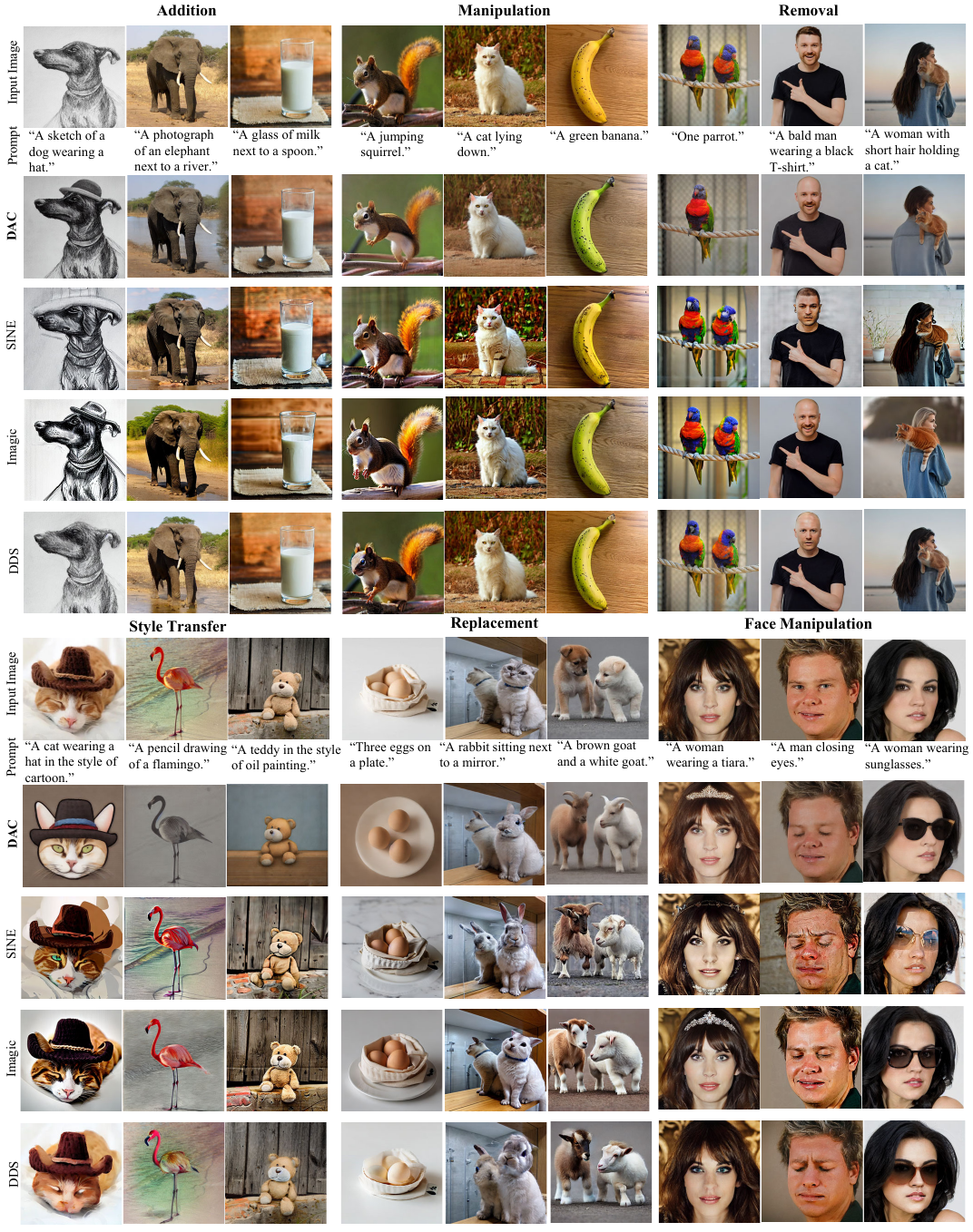}     
    \caption{Comparison of TBIE qualitative examples across the 6 editing types (only prompt $P'$ shown) between our DAC and three SOTAs.}
\label{fig:comparisions2}
\end{figure*}

\begin{figure*}[t]
    \centering
    \includegraphics[width=0.85\textwidth]{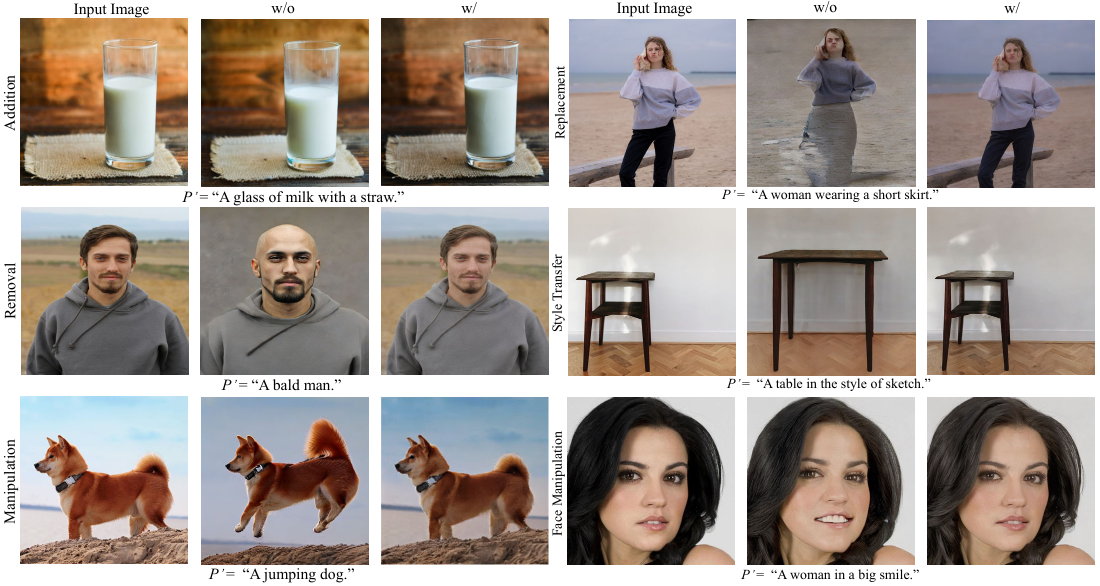}  
    \vspace{-3mm}
    \caption{Ablation on UNet w/o and w/ Conv and FFN LoRA.}
\label{fig:supp_lora}
\vspace{-3mm}
\end{figure*}

\begin{figure*}[t]
    \centering
    \includegraphics[width=0.85\textwidth]{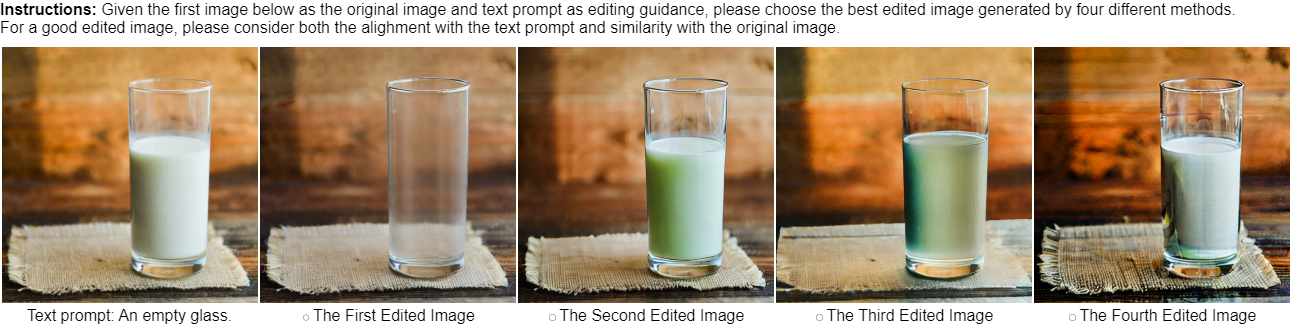}  
     \vspace{-2mm}
    \caption{User study screenshot for one example.}
\label{fig:supp_amt}
\vspace{-3mm}
\end{figure*}

\begin{figure*}[!htp]
    \centering
    \includegraphics[width=0.85\textwidth]{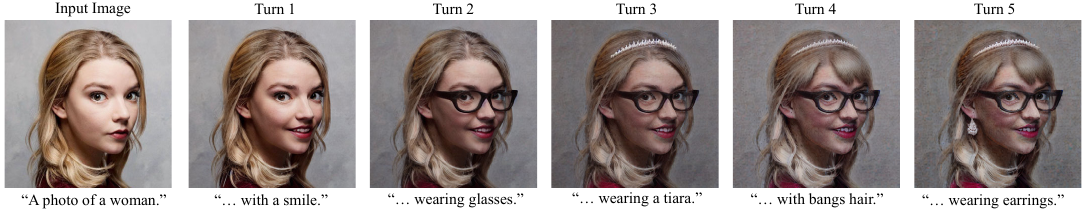} 
    \vspace{-2mm}
    \caption{A qualitative example of muti-turn editing.}
\label{fig:supp_limits}
\vspace{-5mm}
\end{figure*}

\subsection{Qualitative Results}
The examples for qualitative comparisons on InstructPix2Pix dataset~\cite{brooks2023instructpix2pix} are shown in Figure~\ref{fig:supp_comp}. 
For the first example, the text prompt aims to change the yellow taxi to a green one. It could be seen that DAC successfully modifies the color of the taxi while maintaining other components in the input image. 
By contrast, the edited images from the Imagic~\cite{kawar2023imagic} and DDS~\cite{hertz2023delta} are even the same as the source image, inconsistent with the text prompt. 
Considering the second example, all three methods attain effective editing. 
For the third and fourth examples, DAC adds a crown and a woman according to the text prompts separately. The results of DDS~\cite{hertz2023delta} fail to achieve the desired editing although it keeps high fidelity to the input images, thus explaining the best LPIPS score of DDS~\cite{hertz2023delta} in Table \ref{tab:supp_compare}.

Additionally, Figures~\ref{fig:comparisions1} and~\ref{fig:comparisions2} provide extra qualitative comparisons for the six editing types, contrasting our DAC with three state-of-the-art methods.

\section{Ablation Analysis}
\label{b}
\noindent \textbf{Ablation on UNet LoRA.} 
The LoRA structure in DAC is built on all of the attention layers, convolutional layers, and feed-forward (FFN) layers since we observe the underfitting issue if we only apply LoRA on the attention layers of UNet. 
The underfitting issue means that we could modify the image by directly changing the text prompt with $U$.
Figure~\ref{fig:supp_lora} shows the ablation results of w/o and w/ Conv and FFN LoRA in the UNet. 
For $U$ w/o Conv and FFN LoRA, we could get ``A bald man", ``A jumping dog", and ``A woman in a big smile" with the target prompts. However, the fidelity of edited images is lost. For example, the identity of the man even changes in the removal editing of Figure~\ref{fig:supp_lora}. 
By contrast, with $U$ containing Conv and FFN LoRA, we couldn't alter the image anymore, thus overcoming the underfitting issue.

\section{User Study Details}
\label{c}
We quantitatively evaluate our DAC with an extensive human perceptual evaluation study conducted on AMT.
Concretely, we collected a diverse set of image-prompt pairs, covering all the ``addition", ``manipulation", ``removal", ``style transfer", ``replacement", and ``face manipulation" editing operations. Each operation includes 9 different prompt-image pairs, thus constituting 54 examples in total (i.e., examples in Figure \ref{fig:visual_comparisions} in the main paper, Figures \ref{fig:comparisions1}, and \ref{fig:comparisions2}). The number of AMT participants is 110 and for each evaluator, 54 examples are shown. Moreover, one example consists of a source image, a target prompt, and 4 edited images by DAC, DDS, SINE, and Imagic, which were randomly listed. The user study screenshot for one example is depicted in Figure \ref{fig:supp_amt}. We listed instructions of our editing evaluation for evaluators. Note that we emphasize a good edited image should fulfill both the alignment with the text prompt and similarity with the original image.

\section{Limitation}
\label{d}
 In Figure \ref{fig:supp_limits}, we provide a qualitative example of multi-turn editing. As the turn increases, our DAC achieves successful editing aligning with the text prompts while the image quality gradually declines. It is caused by the information loss in Abduction-1 as illustrated in Figure \ref{fig:ablation_ab1} and Ablation on Abduction-1. Although we could complete such loss by incorporating another abduction in Abduction-1, it may be time-consuming. To solve the image quality degradation in multi-turn editing, we need to explore time-efficient fine-tuning (e.g., Fast Diffusion Model \cite{wu2023fast}) for the abduction process in DAC. We leave it to our future work.

\end{document}